\documentclass[twocolumn, switch]{article} 

\usepackage{preprint}

\usepackage{amsmath, amsthm, amssymb, amsfonts}

\usepackage[numbers,square]{natbib}
\usepackage{lettrine}
\usepackage[utf8]{inputenc}	
\usepackage[T1]{fontenc}	
\usepackage{xcolor}		
\usepackage[colorlinks = true,
            linkcolor = purple,
            urlcolor  = blue,
            citecolor = cyan,
            anchorcolor = black]{hyperref}	
\usepackage{booktabs} 		
\usepackage{nicefrac}		
\usepackage{microtype}		
\usepackage{lineno}		
\usepackage{float}			

\usepackage{lipsum}		

\usepackage{newfloat}
\DeclareFloatingEnvironment[name={Supplementary Figure}]{suppfigure}
\usepackage{sidecap}
\sidecaptionvpos{figure}{c}

\usepackage{titlesec}
\titlespacing\section{0pt}{12pt plus 3pt minus 3pt}{1pt plus 1pt minus 1pt}
\titlespacing\subsection{0pt}{10pt plus 3pt minus 3pt}{1pt plus 1pt minus 1pt}
\titlespacing\subsubsection{0pt}{8pt plus 3pt minus 3pt}{1pt plus 1pt minus 1pt}

\title{Deep Convolutional Neural Networks in the Face of Caricature: Identity and Image Revealed}

\usepackage{xwatermark}
\newwatermark[firstpage, color=gray!90, angle=0, scale=0.28, xpos=0in, ypos=-5in]{*correspondence: \texttt{matthew.hill@utdallas.edu}}

\usepackage{authblk}

\author[1\thanks{\tt{matthew.hill@utdallas.edu}}]{Matthew Q. Hill}
\author[1]{Connor J. Parde}
\author[2]{Carlos D. Castillo}
\author[1]{Y. Ivette Colon}
\author[2]{Rajeev Ranjan}
\author[2]{Jun-Cheng Chen}
\author[3]{Volker Blanz}
\author[1]{Alice J. O'Toole}
\affil[1]{The University of Texas at Dallas, U.S.A.}
\affil[2]{University of Maryland, U.S.A.}
\affil[3]{University of Siegen, Germany}

\begin{document}

\twocolumn[ 
  \begin{@twocolumnfalse} 
  
\maketitle

\vspace{-0.35cm}
\begin{abstract}
Real-world face recognition requires an ability to perceive the unique features of an individual face across multiple, variable images. 
The primate visual system solves the problem of image invariance
 using cascades of neurons that convert images of faces into categorical 
representations of facial identity. Deep convolutional neural networks (DCNNs)
also create generalizable face representations, but with cascades of simulated neurons.
DCNN representations can be examined in a multidimensional ``face space'', 
with identities and image parameters quantified via their projections onto the axes that define the space. We examined the organization of viewpoint, illumination, gender, and identity in this space. We show that the network creates a highly organized, hierarchically nested, face similarity structure in which information about face identity and imaging characteristics coexist. Natural image variation is accommodated in this hierarchy, with face identity nested under gender, illumination nested under identity, and viewpoint nested under illumination. To examine identity, we caricatured faces and found that network identification accuracy increased with caricature level, and{\textemdash}mimicking human perception{\textemdash}a caricatured distortion of a face 
``resembled'' its veridical counterpart. Caricatures improved performance by moving 
the identity away from other identities in the face space and minimizing the effects of illumination and viewpoint.   
Deep networks produce face representations that solve long-standing computational problems in generalized face recognition. They also provide a unitary theoretical framework for reconciling decades of behavioral and neural results that emphasized either the image or the object/face in representations,  
without understanding how a neural code could seamlessly accommodate both.
\end{abstract}
\vspace{0.35cm}

  \end{@twocolumnfalse} 
] 

\let\thefootnote\relax\footnote{{\tiny Author contributions: all authors were involved in conceptualization and design of the methodology of the study. MQH, CJP, CDC, RR, and JC handled software. The original draft was prepared by MQH and AJO. Review and editing were done by MQH, CJP, CDC, YIC, VB, and AJO. Formal analysis, investigation, and visualization were done by MQH and CJP, with validation by MQH, CJP, and CDC. Supervision and funding acquisition were handled by CDC and AJO, with project administration by AJO\\

\noindent University of Maryland has filed a US patent application that covers portions of Network A. RR and CDC are co-inventors on this patent.\par}}


People recognize familiar faces effortlessly across changes in
viewpoint, illumination, facial expression, and  appearance (e.g., glasses,
facial hair). The nature of the visual representation that supports this skill
is unknown, despite decades of research in psychology and neuroscience
\cite{marr1982vision,brunelli1993face,riesenhuber1999hierarchical,bulthoff1992psychophysical,yuille1991deformable}.
Hypotheses about face representations posit alternatively that the primate
visual system reconstructs an object-centered facsimile of a face
\cite{marr1982vision,biederman1987recognition} or that it represents multiple
image-based views of faces
\cite{bulthoff1992psychophysical,riesenhuber1999hierarchical,Poggio1990}. The
former is consistent with the ability of humans to recognize familiar faces
(e.g., friends, family) across a wide range of image and appearance variation.
The latter is consistent with well-established difficulties humans have in
perceiving ``identity constancy'' for unfamiliar faces across variable images
\cite{jenkins2011variability}. Although object-centered and image-based models
have made progress on the problem of generalized face/object recognition,
neither provides a unified account of how the visual system simultaneously
discriminates facial identities while managing (filtering out or encoding) image
and appearance variation.

\begin{figure} [!ht]
	\begin{center}
		\includegraphics[width=2.5in]{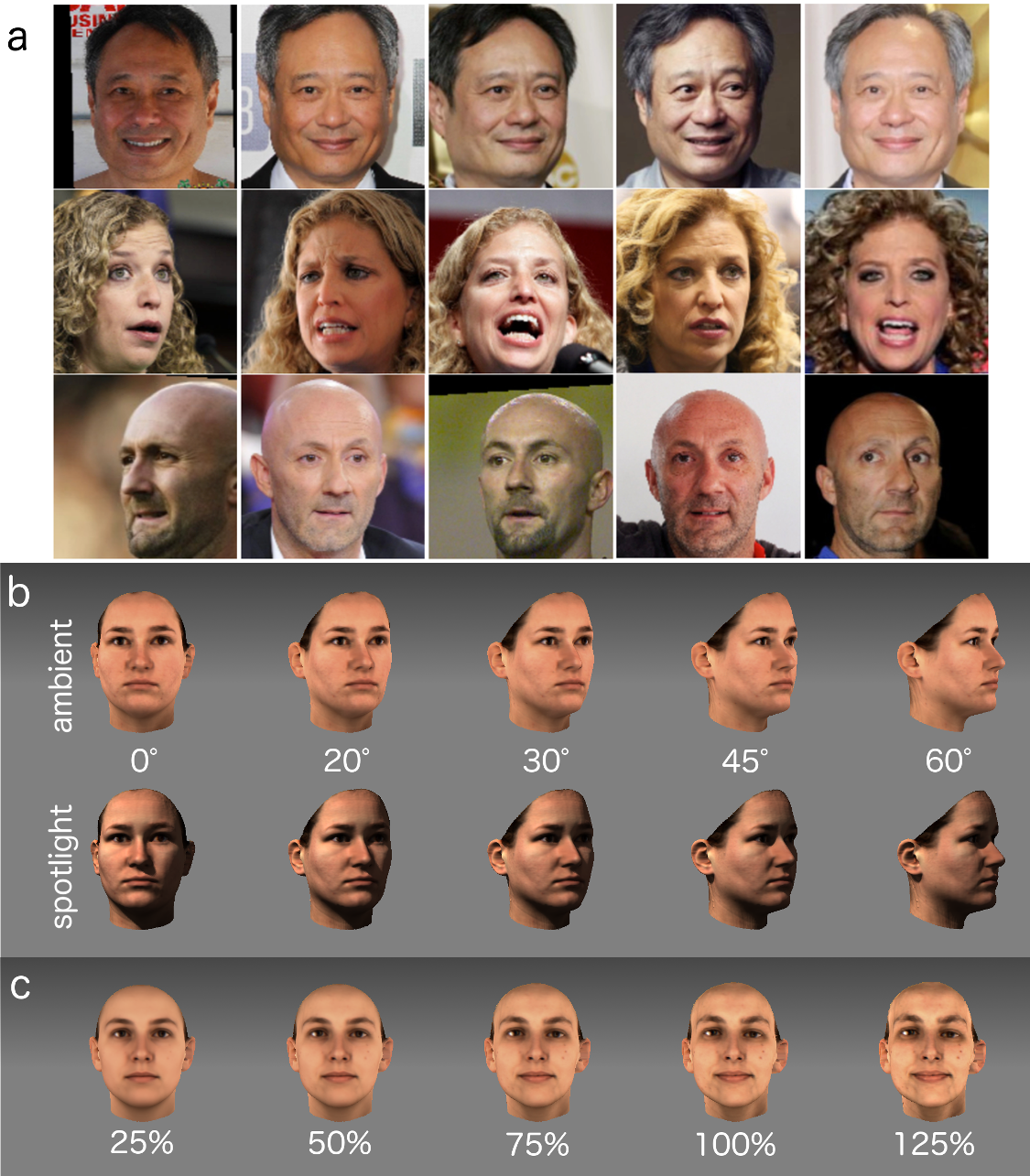}
		\caption{Example of face images used to train and probe the organization
            of imaging characteristics and subject information in the network.
            Training was done on real-world unconstrained face images (a).
            Testing was done on highly controlled laser-scan data varying by
            viewpoint (b, columns), illumination (b, rows), and identity
            strength (c).}
		\label{fig:Example data types3}
	\end{center}
\end{figure}

\begin{figure*}[!ht]
	\begin{center}
		\includegraphics[width=5.50in]{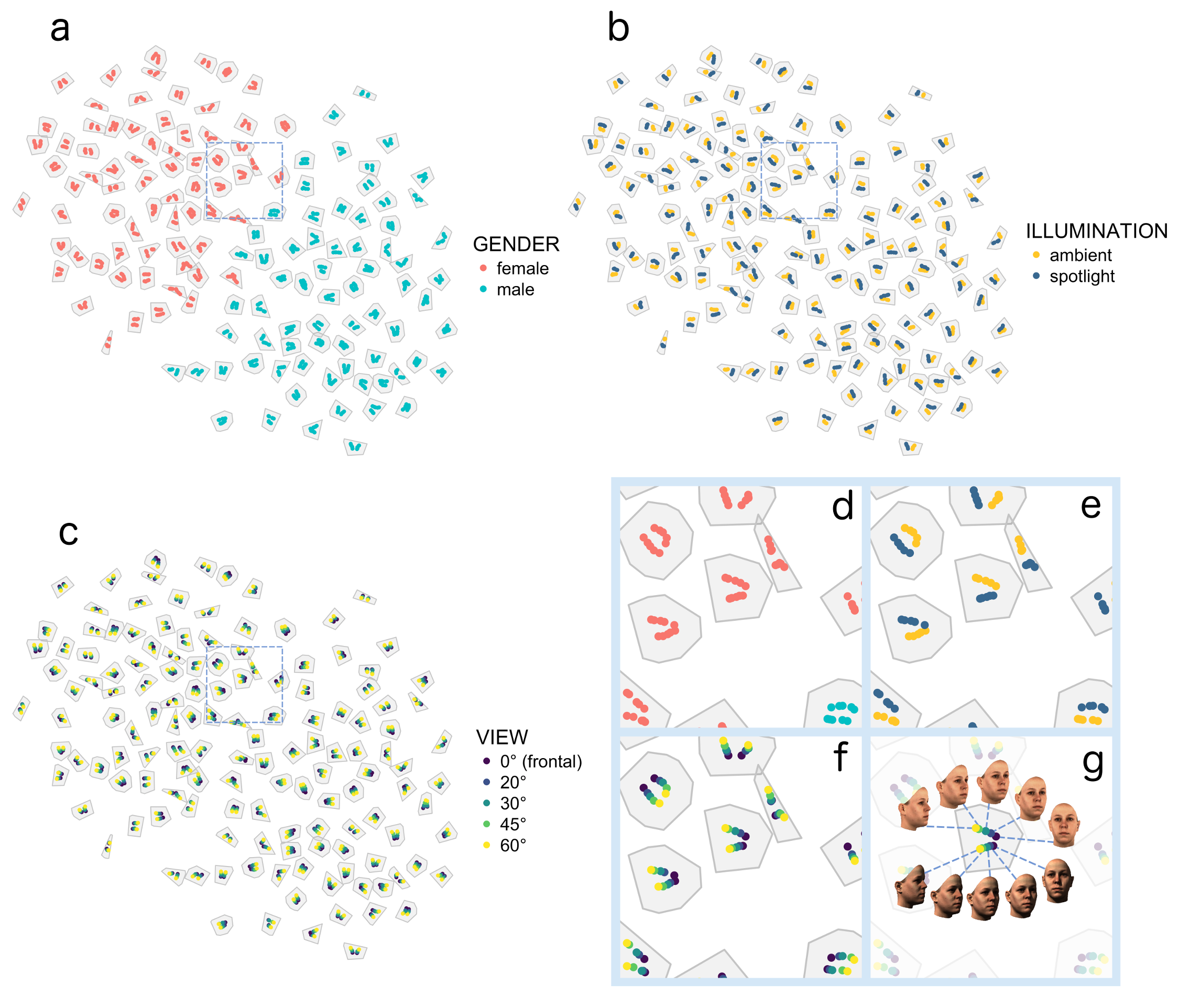}
		\caption{Visualization of top-level DCNN similarity space for all images.
			The network separates identities accurately (gray polygonal borders surround all images of each identity).
			The space is divided into male and female sections (a, d). Illumination conditions  subdivide within identity groupings (b,e).
			Viewpoint varies sequentially within illumination clusters (c, f).
			Dotted-line boxes (a{\textendash}c) show area covered by zoomed-in sections (d{\textendash}g).
		}
		\label{fig:Accuracy over all groups}
	\end{center}
\end{figure*}

Computational models, developed in parallel to the psychological and neural
theories, illustrate clearly the benefits and pitfalls of object-centered and
image-based face representations. In early image-based models, principal
components analysis (PCA) was applied to sets of face images
\cite{turk1991eigenfaces} to create a {\it face space}
\cite{valentine1991unified}. This model accounts for behavioral findings of a
recognition cost for unfamiliar faces when imaging conditions change between
learning and test   \cite{troje1996face}. It also provides insight into the
gender \cite{o1993low}, race \cite{o1994structural}, features \cite{nestor2016feature}, and identity
\cite{o1993low,o1988physical} information  in face images. However, image-based
PCA works only when the learned and  test images are taken under similar
conditions (e.g., viewpoint). Thus, it fails to account for the robust nature of
human recognition of familiar faces.

The failings of image-based models led to the development of 3D morphable models
\cite{blanz1999morphable}, which represent {\it faces} rather than {\it images
		of faces}. These models operate on densely sampled shape and pigmentation
information from laser scans of faces. As with the image-based models, a face
space is created by applying PCA to sets of faces. In this space, individual
identities are defined as trajectories that radiate out from the average face.
As a face moves away from the average along its {\it identity trajectory}, it
becomes increasingly distinctive, changing from anti-caricature to veridical,
and then to caricature. The paradox of caricatures is that they portray a good
likeness of a person with a distorted image. Morphable models implement a
prototype theory of face recognition
\cite{valentine1991unified}
and account for caricature perception \cite{Rhodes1987}. They fail as a model of
human recognition, because there is no mechanism for a face representation to
improve as the face becomes familiar through exposure to more (and more diverse)
images \cite{Cavazos2018,Dowsett2016,Ritchie2017}.

\begin{figure*}[!ht]
	\begin{center}
		\includegraphics[width=6.5in]{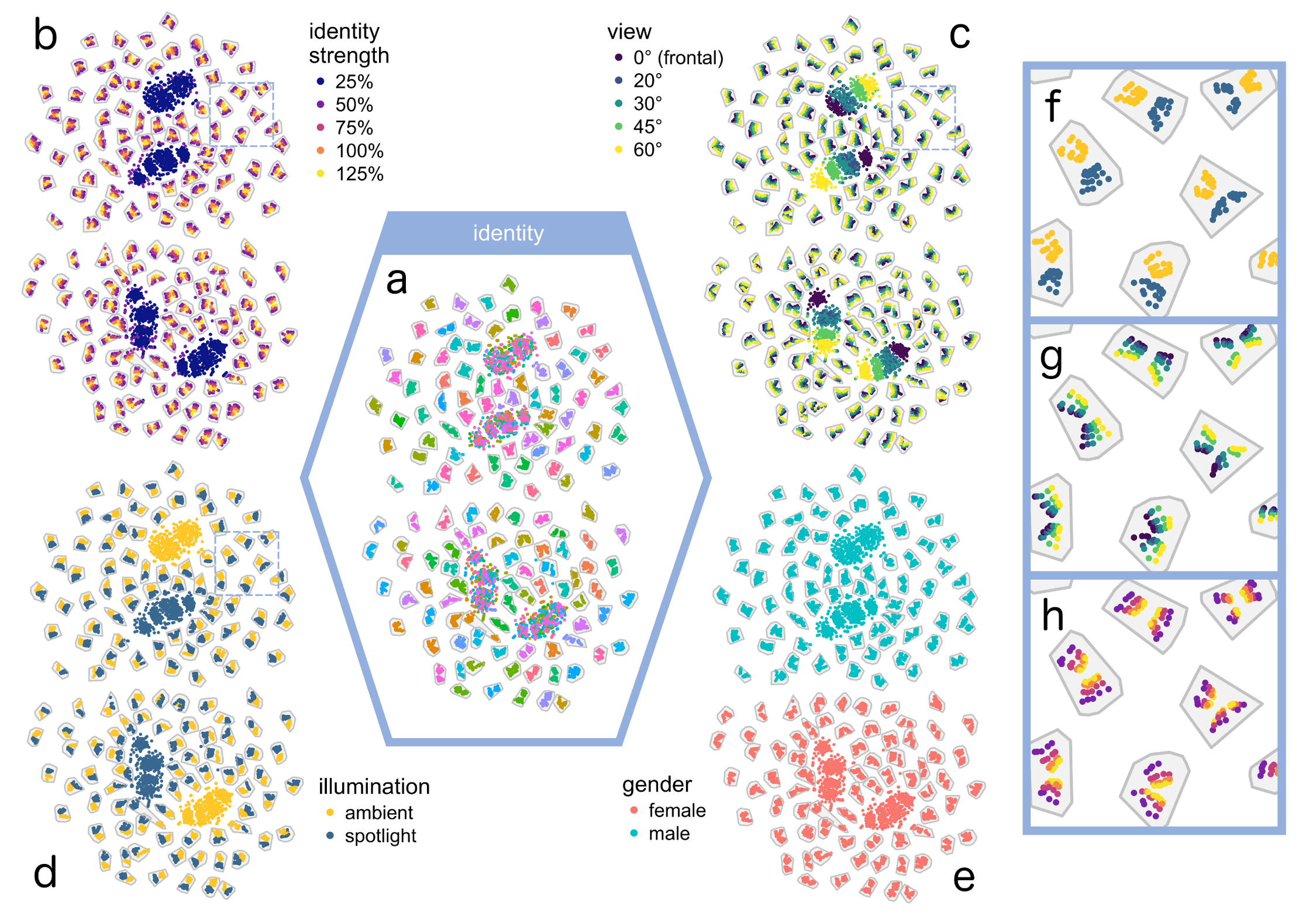}
		\caption{Visualization of top-level similarity space with identity
            strength variation. Mixed-identity clusters appear in addition to
            identity-constant clusters (a), with these mixed regions containing
            weak identity strength images (b). Each identity-mixed cluster
            contains images of a single viewpoint (c), nested within a single
            illumination condition (d),  within a gender group (e). Zoomed-in
            sections (f{\textendash}h) show that, within an identity cluster, images divide
            by illumination conditions (f); viewpoints divide with caricature
            levels arranged in string-shapes (g); caricatures  fall in the
            center of the identity cluster (h). Gray polygons contain all images
            of an identity where identity strength is $\geq$ 75\%.}
		\label{fig:Accuracy over all groups2}
	\end{center}
\end{figure*}

\begin{figure}[!ht]
	\begin{center}
		\includegraphics[width=3in]{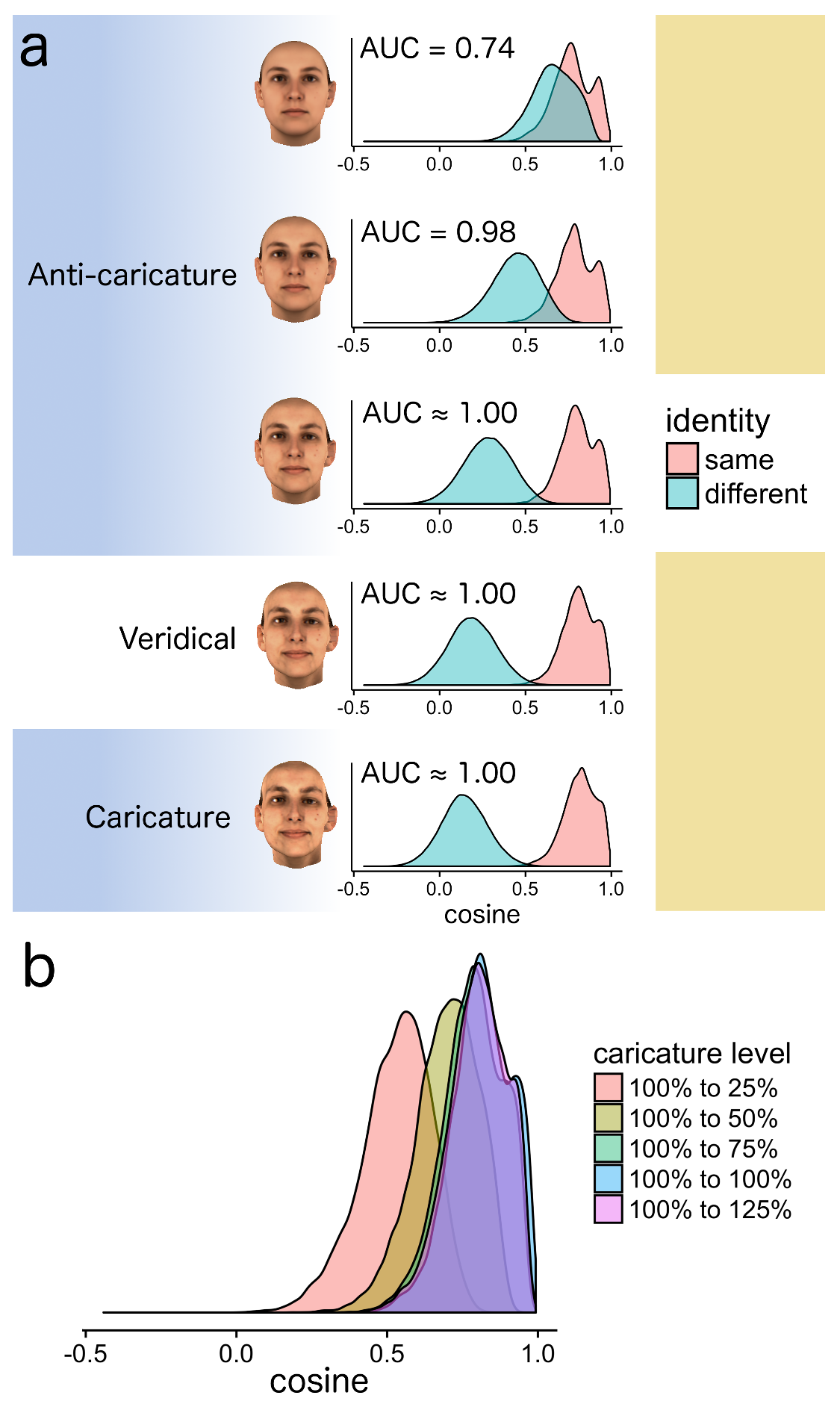}
		\caption{(a) Image-pair similarity score distributions show that accuracy
            increases with caricature level.  This is due to the greater
            dissimilarity of caricatures to other identities (leftward drift of
            the different-identity distribution). (b) Similarity distributions
            across caricature levels show that the veridical is similar to the
            caricature and the 75\% anti-caricature, but not to weak
            anti-caricatures ($\leq$ 50\%).}
		\label{fig:Example data types2}
	\end{center}
\end{figure}

Deep convolutional neural networks are now the state-of-the-art in machine-based
face recognition, because they can generalize identity across  variable images
\cite{sun2014deep,taigman2014deepface,sankaranarayanan2016triplet,schroff2015facenet,chen2015end,ranjan2017all}.
These networks are modeled after the primate visual system
\cite{fukushima1988neocognitron,krizhevsky2012imagenet} and consist of multiple
layers of simulated neurons that perform nonlinear convolution and pooling
operations. DCNN representations expand in early layers of the network, but are
compressed in the top layers through a bottle-neck of neurons. The
representation of facial identity that emerges at the final layer of a DCNN is
compact and can operate robustly over changes in image parameters (e.g.,
viewpoint) and appearance.

DCNN face representations have characteristics of both object-centered and
image-based codes. Similar to object-centered models, they represent identity
with an image-invariant code.  Similar to image-based models, DCNN
representations retain information about the images they process
\cite{parde2017face,o2018face}. Specifically, features from the top-layer of
DCNNs trained for face recognition support reliable linear read-out of the
viewpoint (yaw in degrees, pitch as on-center versus
up/down) of the input image \cite{parde2017face}.

Deep networks offer a proof-of-principle that a robust and general coding of
high level visual information can co-exist with instance-based codes that retain
characteristics of the imaging conditions. But how do DCNN codes accomplish the
balancing act of accommodating facial identity and image information in a
unitary representation? It has been difficult to directly address this question because
viewpoint, illumination, and the number/quality
of images for each identity are not controlled in datasets typically used to
train DCNNs. To
overcome this challenge, we probed a network trained with ``in-the-wild'' face
images using an ``in-the-lab'' dataset. Specifically, we used highly controlled
laser scans of faces to examine how DCNNs represent faces in terms of their
subject parameters (identity and gender) and image characteristics (viewpoint
and illumination). To probe the nature of the identity representation in these
networks, we manipulated the strength of identity information in a face with caricatures.
The results show that DCNNs produce a remarkably organized
representation of faces that is consistent with human perception of face
identity across variable  images and across caricature ``distortions''.

\begin{figure*}
	\begin{center}
		\includegraphics[width=6.75in]{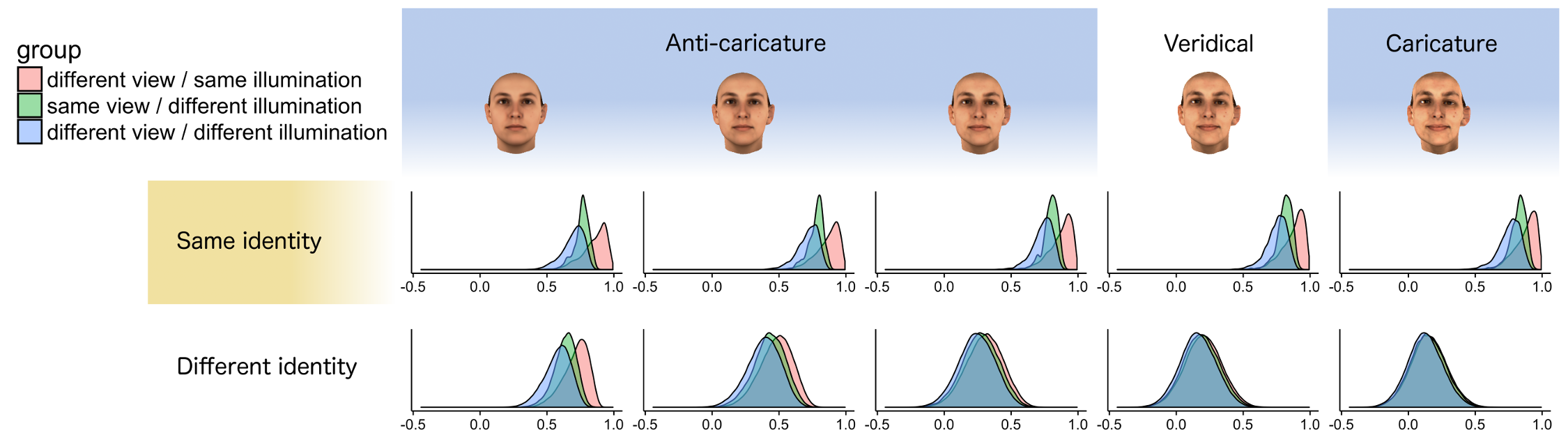}
		\caption{Density curves of face image-pair cosine similarity scores.
            Overlap between same-identity (top row) and different-identity
            (bottom row) distributions decreases as identity strength increases.
            Within same-identity distributions, viewpoint and illumination
            differences are visible at all caricature levels as peaks in the
            distributions. These peaks are visible in the different-identity
            distributions only for weak identity strengths.}
		\label{fig:Example data types}
	\end{center}
\end{figure*}

\section*{Results }

\subsection*{Face Space Visualization}

We examined the organization of imaging characteristics and subject variables in
the DCNN top-layer face representation using a {\it face space} framework
\cite{valentine1991unified, leopold2001prototype}. In this framework, the
distance between points in the space reflects the similarity of face images as
``perceived'' by the top layer of the DCNN. We report data on a 101-layered face
identification DCNN \cite{ranjan2017all} trained with 5,714,444 in-the-wild
images (see Fig. 1A) of 58,020 identities. The top-layer output of the network is a
512-element face representation. Images were created from laser-scans of 70 male
and 70 female heads registered to a parametric 3D face model
\cite{blanz1999morphable}. Each face was rendered from five viewpoints (yaw:
$0^{\circ}$ [frontal], $20^{\circ}$, $30^{\circ}$, $45^{\circ}$, $60^{\circ}$
[left profile]) under two illumination conditions (ambient, directional
spotlight). This produced 1,400 images (Fig. 1B), which we processed through the
DCNN to produce a top-layer representation for each image. To examine the
structure and information content of the face space that emerges at the top
layer of units in DCNNs, we visualized the face-space representations of the
1,400 images using $t$-distributed Stochastic Neighbor Embedding ($t$-SNE)
\cite{maaten2008visualizing,van2014accelerating}.

Figure 2 shows the hierarchical organization of the face space with respect to
gender, identity, illumination, and viewpoint. Identities were separated with
high accuracy (Area Under the Curve [AUC] $\approx$ 1), indicating that the DCNN
recognizes faces across substantial variability in viewpoint and illumination.
The space is separated roughly into two clusters by gender (Fig. 2A). Within
each identity cluster, face images from the two illumination conditions separate
into sub-clusters (Fig. 2B, E). Within each illumination sub-cluster, images are
arranged systematically  by viewpoint, like beads on a chain (Fig. 2C, F). This
demonstrates a highly organized representation of image information in a robust
identity code.

Next, we quantified the accessibility of gender, illumination, and viewpoint  in
the full high-dimensional  space, using a linear classifier. All three variables
were predicted accurately from the face representations ($p < .001$, in all cases). Viewpoint was detected
with an average error of 6.34$^{\circ}$ ($SD = 4.95^{\circ}$), illumination
classification was 95.21\% correct, and gender classification was 98.21\%
correct. This demonstrates accurate linear read-out of image and subject
information from the top-layer face representation.

\subsection*{Identity Strength in the Face Space}

To examine facial distinctiveness, we used the 3D head model to generate morphs
that varied in the strength of the identity information in the face
\cite{leopold2001prototype}. Following the identity trajectory,  each face was
morphed from a caricature (high identity strength) to a near-average face (low
identity strength) in four equal steps. This yielded five versions of each face
(125\% [caricature]; 100\% [veridical]; 75\%, 50\%, 25\% [anti-caricature]). The
addition of identity strength increased the dataset to 7,000 images (Fig. 1C).

Figure 3 shows the $t$-SNE face space with the inclusion of identity strength
variation. It shows that faces with weak identity information are grouped
according to other variables (gender, view, illumination). Specifically, Fig. 3A
shows that mixed-identity clusters are scattered among correctly clustered
identities,  and Fig. 3B shows that mixed-identity regions contain only faces
with weak identity strength. Each identity-mixed cluster contains images of a
single viewpoint (Fig. 3C), nested within a single illumination condition (Fig.
3D), and within a gender group (Fig. 3E). Zoomed-in sections (F{\textendash}H) show that
within an identity cluster, images divide by illumination conditions (Fig. 3F).
Viewpoints also divide, with caricature levels arranged in string-like groups
(Fig. 3G). Caricatures are centered in identity clusters (Fig. 3H),
showing that same-identity caricatures cluster more closely over image variation than veridicals and anti-caricatures
 (see also SI).

\subsection*{Caricature and Identity}

Face identification amounts to a decision of whether two images depict the same
or different identities. This decision is based on the cosine similarity between
the top-layer representations of the two images (higher similarities suggest the
same identity). Accuracy can be visualized using the similarity distributions
for same- and different-identity image pairs (wider separation indicates higher
accuracy).

Figure 4A shows that caricaturing improves the network's identification accuracy
by increasing the ``perceptual'' contrast between faces as caricature level
increases (leftward drift of different-identity distribution). Caricaturing does
not appreciably move the same-identity distribution. However, consistent with
its effects on minimizing the impact of imaging parameters (Fig. 3H), the range
of similarity values in this distribution compresses as caricature level
increases (see SI). Next, we asked whether the DCNN ``sees'' the caricature as
the same identity as its corresponding veridical face. Figure 4B indicates that
it does. We looked at the similarity between veridicals and their corresponding
images across caricature level. The network perceives 75\% anti-caricatures and
caricatures as nearly equivalent to veridicals (Fig. 4B). The 25\% and 50\%
anti-caricatures are less similar to their veridical faces.

Caricaturing, therefore, affects DCNN perception by exaggerating a face's unique
identity information relative to other faces in the population without impairing
identity perception.

\subsection*{Caricature and Image Conditions: Viewpoint and Illumination}

How does image-based information interact with identity constancy? Figure 5
shows  that imaging conditions affect the DCNN's perception of face similarity.
Changes in viewpoint and/or illumination can be seen as peaks in the similarity score
distributions for same-identity pairs (top row), at all levels of caricature.
For higher identity strengths  ($\geq$ 75\%), different-identity
distributions (bottom-row) separate visibly from same-identity
distributions, and the salience of image-based similarity is attenuated.  This
shows that identity{\textemdash}not imaging condition{\textemdash}is the primary determiner of
dissimilarity for different-identity pairs. Imaging condition effects reappear
with weak identity strengths ($\leq$ 50\%). These near-average faces approach a
single (average) identity that varies only by imaging condition. Therefore,
similarity in the DCNN encompasses both identity and viewing conditions, but on
a different scale. Identity contributes far more than image conditions.

\section*{Discussion}

Deep networks accomplish the balancing act of accommodating facial identity and
image information in a unitary representation by generating an elegantly
organized face similarity space. To understand this organization, it is useful
to distinguish between person properties (e.g., identity, gender, race) and
specific image encounters (e.g., viewpoint, illumination). The former, immutable
characteristics, divide the  face space into subspace partitions that are
homogeneous with respect to their defining person characteristics.  The latter,
variable properties, are accommodated within the homogeneous subspaces, yet they
apply to all identities.

Being able to access information about  object/person  properties at multiple
levels of abstraction is a computational goal of  a visual categorization system
\cite{grill2014functional}. In psychological terms, the topology of the DCNN
space  organizes faces to allow easy access to person properties at different
levels of abstraction. The  space itself defines a basic-level category of
faces. The position of an image in the face space indicates a subordinate
gender-category, and the position in this gender category specifies an exemplar
category of identity \cite{rosch1976basic,grill2014functional}.

To access the fundamental person property of identity, the DCNN must code the
uniqueness of a face across variable image conditions.  The robust nature of
this unique identity information in a DCNN is inherited from the topology of its
similarity space, which is highly non-linear with respect to image properties.
Two widely different images (e.g, frontal versus profile) are coded as similar,
because the network represents identity categorically.  The use of caricaturing
to probe the organization of the face space provides a unique vantage point for
seeing how identity and image information interact in a DCNN. Caricaturing
affects the DCNN performance because it operates both within individual identity
clusters and at the level of face populations. Within identity clusters,
caricatured faces minimize the influence of imaging parameters. At the
population level, caricaturing increases the separation between face identities
in the space, making them all less confusable.

From a psychological perspective, the DCNN's combined representation of identity
and image encounters provides a unified account of behavioral effects seen
previously as evidence for exclusively image-based {\it or} object-centered
theories of face processing. DCNN representations are compatible with a face
recognition cost for changes in image parameters between learning and testing.
They are also compatible with effects of face distinctiveness relative to a
population. The general accord between behavioral results and  deep network
representations, combined with the network's ability to produce a robust
representation of identity, makes DCNNs a plausible model of human face
processing. The present work with in-the-lab images points to the possibility of
addressing how ``familiarity'' with a face, via exposure to in-the-wild images,
might alter the capacity of the face representation to generalize recognition
even further. There are multiple stages of DCNN training that can be targeted in
this endeavor \cite{o2018face}.

From a neuroscience perspective, DCNN representations reconcile the seemingly
 paradoxical
nature of ventral temporal cortex organization as both object-categorical {\it
and} reflective of low level image properties, e.g., viewpoint
\cite{kietzmann2015occipital,natu2010dissociable}, illumination
\cite{grill1999differential}, size \cite{yue2010lower}, and position
\cite{kay2015attention}. For the former, structure exists in the organization of
person properties in subspaces. For the latter, structure within identity
subspaces is duplicated across identities to index image properties.

From a computational perspective, converting a representation in the image
domain to one that operates in a categorical domain, does not necessarily entail
information loss. Instead it can be achieved by reorganizing the space. Although
much of what we see of this organization here is sufficiently salient to be
visualized in two dimensions, the full representation in the high dimensional
space drives these effects (and our computations). If the goal of a visual
system is to reorganize the representational codes to ``untangle'' information
that is non-linear in the image domain \cite{dicarlo2007untangling}, then the
data configurations we arrive at here may offer a first look at how cascades of
neural-like computations can represent face identity robustly with limited loss
of image context.

\footnotesize
\section*{Methods}

	\subsection*{Networks}
	To test the stability of the face space across network architectures and
	training data, we performed these simulations on two face identification
	DCNNs: Network A \cite{bansal2017s,ranjan2018crystal} (main text), and
	Network B \cite{chen2016unconstrained}. Network A is a ResNet-based DCNN
	trained with the Universe dataset \cite{bansal2017s,ranjan2018crystal},
	which is a mixture of three datasets (UMDFaces \cite{bansal2017umdfaces},
	UMDVideos \cite{bansal2017s}, and MS1M \cite{guo2016ms}). It includes images
	and video frames acquired in extremely challenging, in-the-wild conditions
	(pose, illumination, etc.). We used the ResNet-101
	\cite{wen2016discriminative} architecture with the Crystal Loss (L2 Softmax)
	 loss function for training \cite{ranjan2018crystal}. ResNet-101 consists of 101
	layers organized with skip connections that retain error signal strength to
	leverage very deep CNN architectures. Scale factor $\alpha$ was set to 50.
	The final layer of the fully-trained network was removed and the penultimate
	layer (512 features) was used as the identity descriptor. Once the training
	is complete, this penultimate layer is considered the ``top layer.'' Network
	B has 15 convolution and pooling layers, a dropout layer, and a fully
	connected top layer that outputs a 320-dimensional identity descriptor.
	Network B was trained using a softmax loss function on the CASIA-WebFace
	dataset (494,414 images of 10,575 identities that vary widely in
	illumination, viewpoint, and overall quality [blur, facial occlusion, etc.]).

	\subsection*{Morphing Stimuli}

	Stimuli were made from 3D laser scans with densely sampled shape and
	reflectance data from faces.  These scans were put into point-by-point
	correspondence with an average face. In this format, a face is described
	as a deformation field from the average face, in shape $[ \delta  x, \delta y, \delta z ]$
	and reflectance $[ \delta r, \delta g, \delta b ]$. Identity strength was manipulated
	by multiplying the face representation by a scalar value, $s$, such that $s > 1$ 
	produces a caricature; and $0 < s < 1$ produces an anti-caricature.

	\subsection*{Visualization}
	Face space visualizations were done with $t$-SNE, a non-linear
	dimensionality reduction technique that uses gradient descent to preserve
	the distance between each point in a high-dimensional space, while reducing
	the number of dimensions \cite{maaten2008visualizing}. DCNNs use the angular
	distance between representations to compare images. To preserve this
	relationship in the space, face representation vectors were normalized to
	unit length before computing the $t$-SNE. We used the Barnes-Hut
	implementation of $t$-SNE \cite{van2014accelerating} with a $\theta$ of 0.5, and perplexity
	coefficients of 30 (Fig. 2) and 100 (Fig. 3). $t$-SNE was used to reduce the
	DCNN's 512-dimensional feature space to a two-dimensional space. However, all
	quantitative analyses were conducted in the full 512-dimensional space.

	\subsection*{Classification}

	Linear discriminant analysis (LDA) was applied to the full-dimensional face
	descriptors to classify gender and illumination. Linear regression with the
	Moore-Penrose pseudo-inverse was used to predict viewpoint. Predictions were
	conducted with identity-level cross-validation. Classifications generated
	from Network B produced results similar to those generated from Network A
	(see SI). Statistical significance of the Network A predictions were
	evaluated with permutation tests. A null distribution was generated from the
	original data matrix by creating random permutations of the column contents.
	Permutations ($n=1000$) were generated for each variable (gender,
	illumination, viewpoint). Resulting distributions were compared to the true
	value from each classification test. All permutation tests proved
	significant at $p < .001$, with no overlap between test value and null
	distribution. Network B produced the same results.

\footnotesize
\section*{Acknowledgements}

Funding: Supported by the Intelligence Advanced Research Projects
	Activity (IARPA). This research is based upon work supported by the Office
	of the Director of National Intelligence (ODNI), Intelligence Advanced
	Research Projects Activity (IARPA), via IARPA R\&D Contract No.
	2014-14071600012. The views and conclusions contained herein are those of
	the authors and should not be interpreted as necessarily representing the
	official policies or endorsements, either expressed or implied, of the ODNI,
	IARPA, or the U.S. Government. The U.S. Government is authorized to
	reproduce and distribute reprints for Governmental purposes notwithstanding
	any copyright annotation thereon.

\normalsize
\bibliography{fs}

\clearpage
\onecolumn

\renewcommand{\thefigure}{S\arabic{figure}}
\setcounter{figure}{0}

\section*{Supplemental Information}

In this supplemental information section we report further details and additional analyses for Network A. We also report replications using Network B for classification analysis and network performance.

\section*{Face Space Visualization}

\subsection*{Classification Replication}

Linear discriminant analysis (LDA) was applied to the full-dimensional face
descriptors to classify gender and illumination. Linear regression with the
Moore-Penrose pseudo-inverse was used to predict viewpoint. Predictions were
conducted with identity-level cross-validation. Classifications generated
from Network B produced results similar to those generated from Network A.

\begin{table}[!htb]
\centering
\caption{\small Classification and regression results for Networks A and B. Percentages (\%) denote classification percent correct, while degrees ($^{\circ}$) denote average prediction error.}
\label{tab:BenchAccuracyAlgorithms}
{\small
\begin{tabular}{llll} \hline
 & Gender & Illumination & Viewpoint (SD) \\ \hline
Network A & 98.21\% & 95.21\% & 6.34$^{\circ}$ (4.95$^{\circ}$) \\
Network B & 90.98\% & 97.44\% & 7.28$^{\circ}$ (5.73$^{\circ}$) \\
\hline

\end{tabular}
}
\end{table}

\subsection*{Permutation}

Statistical significance of Network A predictions were
evaluated with permutation tests. A null distribution was generated from the
original data matrix by creating random permutations of the column contents.
Permutations ($n=1000$) were generated for each variable (gender,
illumination, viewpoint). Resulting distributions were compared to the true
value from each classification test. All permutation tests proved
significant at $p < .001$, with no overlap between test value and null
distribution.

\begin{figure}[!htb]
\begin{center}
\includegraphics[width=5.5in]{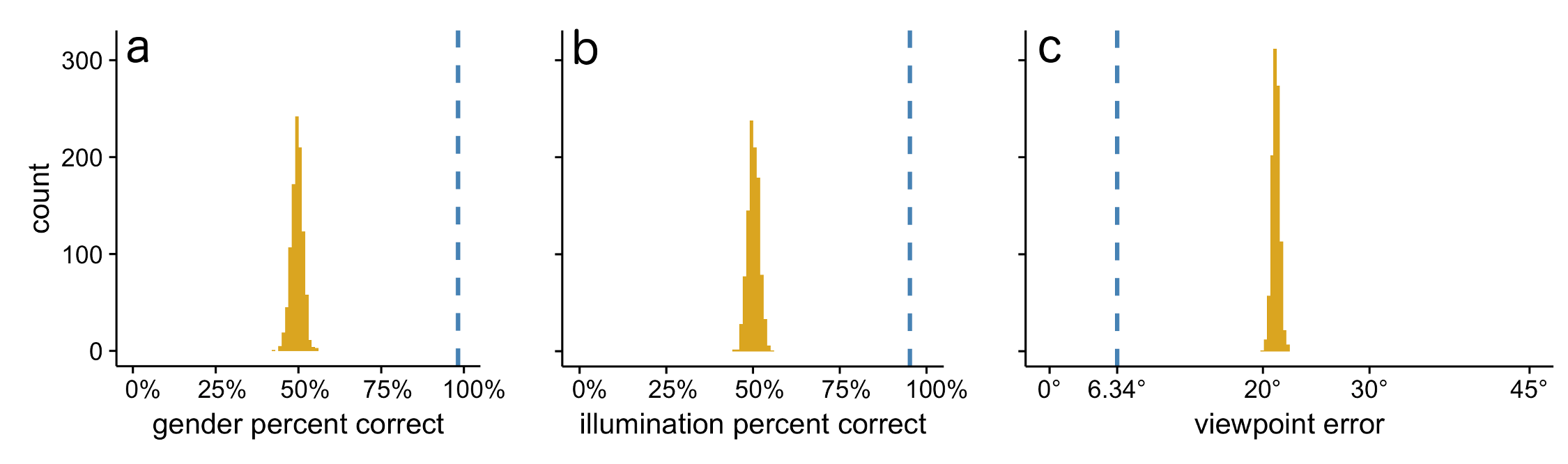} 
\caption{\small Permutation results for Network A show the statistical significance of regression and classification analyses. Yellow histograms show null distributions comprised of 1,000 permutations. Blue dotted lines show test values. Gender showed no overlap between null and test value 98.21 (a), illumination showed no overlap between null and test value 95.21 (b), viewpoint showed no overlap between null and test value 6.34 (c).}
\label{fig:perm}
\end{center}
\end{figure}

\section*{Caricature and Identity}

\subsection*{Network Performance}

We measured the network's identification performance using area under the ROC curve (AUC). It is common in much of the face recognition literature on computational models to
construct the distribution of different-identity image pairs from all possible pairs of images of different identities. Strictly speaking, however,
to measure face identification performance, it is more conservative to 
control for factors other than identity (e.g., gender) that could underlie dissimilarity between faces. Therefore, the
AUCs reported in Figure 4A (main text) include only same-gender image pairs in the different-identity distribution. 

Comparable data to those reported in Figure 4A for Network B are displayed in Table S2.

\begin{table}[!htb]
\centering
\caption{\small AUC scores of each DCNN at each identity strength level.}
\label{tab:BenchAccuracyAlgorithms}
{\small
\begin{tabular}{llllll} \hline
Identity Strength & 25\% & 50\% & 75\% & 100\% & 125\% \\ \hline
Network A (AUC) & 0.735 & 0.979 & 0.999 & 1.000 & 1.000 \\
Network B (AUC) & 0.673 & 0.911 & 0.983 & 0.996 & 0.998 \\
\hline

\end{tabular}
}
\end{table}

\subsection*{Identity Constancy and Caricature}

Figure S2 shows that the similarity scores for same-identity pairs increase marginally with caricature level. This demonstrates a small graded improvement 
in identity constancy over changes in view and illumination as identity strength increases.

\begin{figure}[!htb]
\begin{center}
\includegraphics[width=4in]{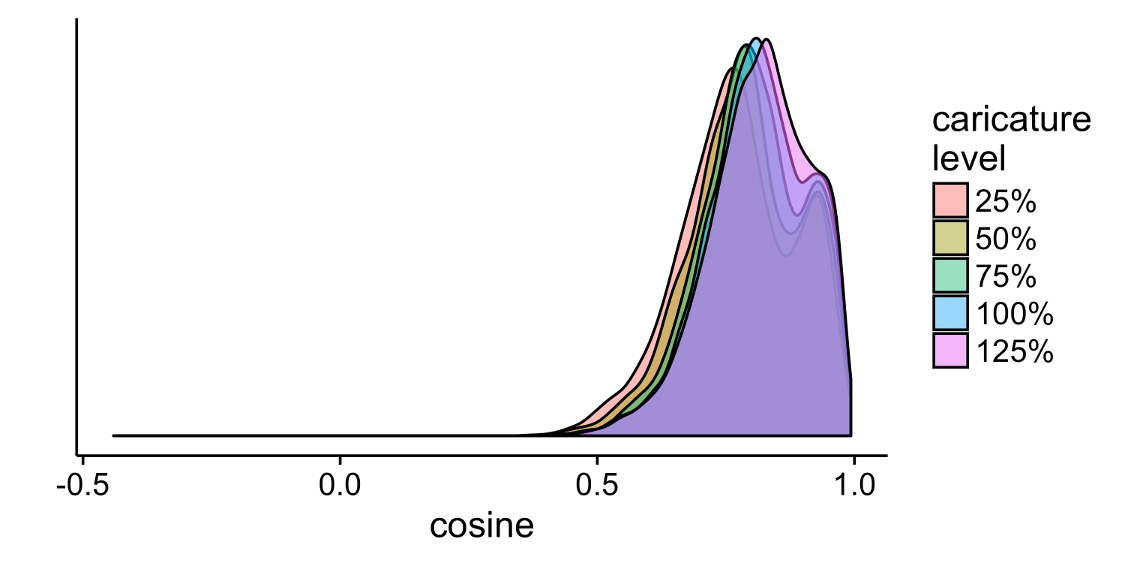} 
\caption{\small Distributions of same-identity similarity scores show that identity constancy increases with caricature level. As the caricature level increases, the range of similarity values from matched-identity image comparisons compresses towards 1.}
\label{fig:matchmorph}
\end{center}
\end{figure}

\subsection*{Imaging Conditions and Caricature}

Figure S3 shows a complete breakdown of the effects of viewpoint and illumination on similarity scores for same- and different-identity image pairs across caricature levels.
This complements Figure 4A in the main text that shows the complete dataset without dividing by the type (viewpoint, illumination, illumination and viewpoint) of image mismatch. 

\begin{figure}[!htb]
\begin{center}
\includegraphics[width=4.4in]{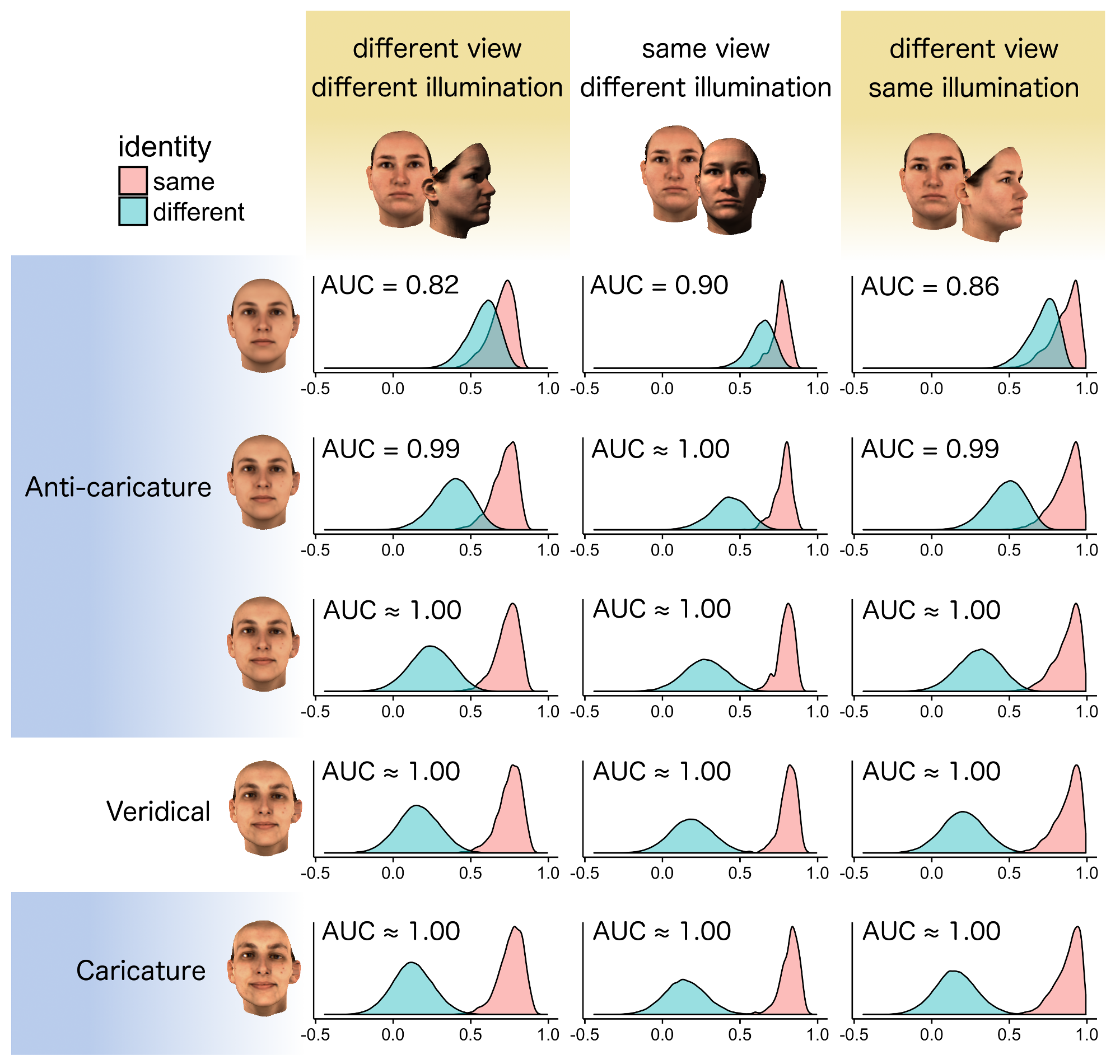} 
\caption{\small Image-pair similarity distributions show that identification accuracy increases with caricature level. This effect is consistent across image comparisons including changes in both viewpoint and illumination, changes in only illumination, and changes in only viewpoint. This increase in identification accuracy is the result of a leftward drift of the non-match distribution and demonstrates that caricaturing benefits performance by accentuating the image features that make two different identities look less like one another.}
\label{fig:bigfig}
\end{center}
\end{figure}

\end{document}